# Lecturer Performance System Using Neural Network with Particle Swarm Optimization

**Abstract:**

The field of analyzing performance is very important and sensitive in particular when it is related to the performance of lecturers in academic institutions. Locating the weak points of lecturers through a system that provides an early warning to notify or reward the lecturers with warned or punished notices will help them to improve their weaknesses, leads to a better quality in the institutions. The current system has major issues in the higher education at Salahaddin University-Erbil (SUE) in Kurdistan-Iraq. These issues are: first, the assessment of lecturers' activities is conducted traditionally via the Quality Assurance Teams at different departments and colleges at the university, second, the outcomes in some cases of lecturers' performance provoke a low level of acceptance among lectures, as these cases are reflected and viewed by some academic communities as unfair cases, and finally, the current system is not accurate and vigorous. In this paper, Particle Swarm Optimization Neural Network is used to assess performance of lecturers in more fruitful way and also to enhance the accuracy of recognition system. Different real and novel data sets are collected from SUE. The prepared datasets preprocessed and important features are then fed as input source to the training and testing phases. Particle Swarm Optimization is used to find the best weights and biases in the training phase of the neural network. The best accuracy rate obtained in the test phase is 98.28 %.

## 1. INTRODUCTION

Assessing the role performance of institution lecturers has a great impact on academic institutions, as it takes control of the educational quality, promising not just the expert skills of their teaching personnel, but additionally, the quality of the teaching and learning. The lecturer and teaching performance analysis and evaluation are understood to be a local assessment that the academic institutions carry out on their academic teaching personnel to accomplish the teaching aims and goals of the institution [1]. Ministry of higher education of Kurdistan Regional Government (KRG) of Iraq owns its quality assurance department which evaluates and observes the activities and performance of quality of teaching for lecturers. The current system is conventionally performed, and it has major problems such as the assessment of lecturers' activities is carried out manually via the Quality Assurance office at the university which produces results that spread low levels of objectivity among lectures, furthermore, the current system is not accurate and vigorous and it is composed of evaluating lecturers against three main criteria which each of them individually includes special features, these criteria are namely; Student Feedback (FB), Lecturers' Portfolio (PRF) and Continuous Academic Development (CAD).

It has been established during the course of this research work that little or no research works were formerly piloted to determine lecturer performance at universities in general, and since classification of lecturer performance system has not been used in Kurdistan region, subsequently, we believe that it is very imperative to use this study as it encounters to regulate the suitable system for solving the responsibilities of lecturer performance.

However, in the previous few years, performance analysis has become an essential playing field within investigators in all profession capacities. The researchers have focused on the lecturer performance application [1, 2] and the student performance application [3-7] which is moderately close to the lecturer performance application. Chaudhari in 2012 proposed a model for evaluating teachers overall performances by using Fuzzy Expert System (FES). The main dataset attributes were student feedback, Result, Students Attendance, Teaching Learning Process activities, Academic Development of Teacher and Other performances. The input data variables are fuzzified into excellent, very good, good, average and poor. Trapezoidal membership function is used to convert the crisp set into the fuzzy set. Fuzzy rules are set to determine input and output membership functions that will be used in inference process [1]. An expert technique is proposed for lecturers' performance evaluation using fuzzy logics [2]. For attribute selection, they extracted a set of 99 attributes from literature review in [3] which are having great impact on assessing lecturers' performance. If else expert system method is built which uses the fuzzy set and membership functions for decision making, and to map linguistic characteristics' of lecturers' performance.

A hybrid approach proposed for classifying student academic performance by using Bayesian probabilities with neural networks to classify students [4], both standard neural network with random weight initialization and a hybrid method composed of Bayesian probability for weight initialization with back propagation feed forward neural network (BPFFNN) are used, then, they compared both methods. As a result, they concluded that BPFFNN performed better than other types of neural network. Data mining techniques were presented for student academic performance. Smooth Support Vector Machine (SSVM) is used to predict student performance, whereas, kernel k-means is used to cluster similar student characteristics in the same group. Clustering was performed to the dataset, then they discovered the logical rules of students' final grade by using J48 decision tree. RBF was used as a type of kernel function for SSVM claiming that RBF produced a good performance and it had a few numbers of parameters [5].

This paper proposes a dynamic and computer based system to SUE so that to make the process of lecturer performance and analysis more productive.

The paper is structured as follows: section two presents theories of algorithm, section three introduces the proposed system in detail, section four describes a methodology involved in this paper and finally, conclusions and future work are presented.

## 2. THEORIES OF ALGORITHM

This paper depended on two algorithms, these are namely; Neural Network with Back Propagation learning algorithm and Particle Swarm Optimization (PSO). Details of these are algorithms are described below:-

### 2.1 Neural Network with Back Propagation

Back propagation is a supervised learning algorithm which is used for training a network. The data is used as input to the network structure and the class labels. Then it is used for measuring the performance of both the training and testing stages of the network. The network input layer firstly is fed with the feature set then the layer neurons spread the data values to the hidden layer neurons. There two calculation stages exist in each hidden layer neuron, which are summing up the values entering it, and then, applying an activation function to the collected values which the result acts as output of that neuron. Starting from applying the input features to the network and finishing with the output values is called an epoch. Weights of the network is then updated continuously after each epoch during the learning stage. Training termination is achieved when either it reaches to the maximum number of epochs or it reaches to an acceptable output value. Finally, test samples are fed to the network in order to evaluate the network performance.

The architecture of neural network trained with Back propagation learning algorithm (BPNN) is composed of an input layer, one or few hidden layers, and an output layer. The BPNN is re-iterative to diminish the error between the real and target outputs. The training session for the BPNN will be as follows [8, 9]:-

The input value for each neuron node can be calculated as:

$$y_j = \sum_{i=1}^{m} w_{ij} x_i \qquad (1)$$

Where *m* is the number of connection lines that enter the neuron *j* and $X_i$ is the output value of *i* neuron from the previous layer. The output of neuron *j* ($y_j$) can be calculated by applying its input value through a specific threshold activation function or by using logistic activation function. Then this output value of neuron *j* will be used as one of the inputs of neuron of the next layer. After computing all the outputs then weights are updated based on the error between the actual output and the desired output. Two techniques are used for weight update and initialization which are Gradient Descent and Particle Swarm optimization. The latter is described in the final part of this section. The weight update of the Gradient Descent is in the form of:

$$\Delta w_{ij} = \zeta \delta_j x_i \qquad (2)$$

Where $\Delta w$ the weight change, $\zeta$ is the learning rate, and $\delta_j$ is the error term of neuron $j$. New weights can be computed recursively at the output neuron to the first hidden layer neuron via:

$$w_{ij}(t+1) = w_{ij}(t) + \zeta \delta_j x_i \qquad (3)$$

Where $w_{ij}(t)$ is the weight from hidden neuron $j$ or from an input to neuron $i$ at time $t$, $x_i$ is either the output of neuron $i$ or is an input, $\zeta$ is the gain term or the learning rate, and $\delta_j$ is an error term for neuron $j$. If neuron $j$ is an output neuron, then

$$\delta_j = y_j(1 - y_j)(d_j - y_j) \qquad (4)$$

Where $d_j$ is the desired output of neuron j and $y_j$ is the actual output. If neuron $j$ is a hidden neuron node, then:

$$\delta_j = x_j(1 - x_j) \sum_k \delta_k w_{jk} \qquad (5)$$

Where $k$ is over all neuron nodes in the layers above neuron $j$. Convergence is sometimes faster if a momentum $\alpha$ term is added and weight changes are smoothed by:

$$w_{ij}(t+1) = w_{ij}(t) + \zeta \delta_j x_i + \alpha(w_{ij}(t) - w_{ij}(t-1)) \qquad (6)$$

## 2.2 Particle Swarm Optimization

Particle Swarm Optimization algorithm was first found by Kennedy and Eberhart in 1995 and it is considered as a nonlinear function algorithm. It is associated with artificial life such as fish schooling and bird flocking in general, in other words, it is more associated with genetic and evolutionary algorithms. Fundamentally, the algorithm applies plain mathematical processes which are employed without difficulties as it has a low-cost computation in speed and memory [10, 11, 12]. The concept of PSO displays rivalry and collaboration among different particles as replacement operation for using common operators in genetic algorithm. A particle or so called a bird is regarded as a possible solution in a particular group. A particle can modify its flying based on its experience and the experience of other particles in the group or so called swarm. The modification is carried out in position and velocity of the particles. The velocity is calculated according to the following equation (7) [10, 11, 12].

$$v_i(t) = v_i(t-1) + c1 * r1 * (bp_i - x_i(t-1)) + c2 * r2 * (bp_{gi} - x_i(t-1)) \qquad (7)$$

where $v_i(t)$ is new velocity of particle $i$, $v_i(t-1)$ is the current velocity of particle $i$, $c1$ and $c2$ are two positive constants, $r1$ and $r2$ are two random numbers between [0,1], $bp_i$ and $bp_{gi}$ are best position or so called solution met by particle and social respectively. The second part of the equation (7) is called cognitive in which a private thinking of the particle is represented, though the third part is called social (global or neighbors) since it represents the collaboration among all particles and the positions can be calculated by equation (8):-

$$x_i(t) = x_i(t-1) + v_i(t) \qquad (8)$$

$x_i(t)$ and $x_i(t-1)$ are the new and previous flying positions of particles respectively. The performance of each particle is obtained according to the pre-defined fitness function which is related to the problem to be solved [11]. The first part of the equation (7) is modified by adding inertia weight $w$ to make a balance between local and global searches as it is shown below in equation (9) [11]:-

$$v_i(t) = w * v_i(t-1) + c1 * r1 * (bp_i - x_i(t-1)) + c2 * r2 * (bp_{gi} - x_i(t-1)) \qquad (9)$$

The following PSO pseudo code provides a better and inclusive understanding about the algorithm, and how the whole thing operates:-

*Begin*
   *For each particle in the group*
     *Randomly initialize particle*
   *END*
   *Do*
     *For each particle in the group*
       *Fitness value is computed for the selected particle*
       *Check if the value of fitness is better than the value of best fitness (bpi) in history*
         *Set the current value as the new bpi for the selected particle*
     *End*

    *Decide on the particle that has the best fitness value among all the particles (as the $bp_{gi}$)*
    *For each particle in the swarm*
      *Calculate particle velocity according to equation (9)*
      *Update particle position according to equation (8)*
    *End*
  *While maximum iterations or minimum error criteria is not reached*
*End*

The algorithm is implemented to determine the best weight and bias set in the training phase of the network, then, they are fed to testing session to find best possible result for the system.

## 3. PROPOSED METHOD

The proposed lecturers' performance enhancement system using particle Swarm Optimization Combined Neural Network (PSOCNN) is used in this paper. The PSOCNN system is an enhanced version of a combined dataset system which has Particle Swarm Optimization (PSO) in the training session. Figure 1, is a flowchart that shows the main processes of PSOCNN. The software is built using java programming language. The combined system in the software is simpler where all the Sub-Dataset are combined and trained with PSOCNN. The next step is to test the network and finally evaluate the results. The PSOCNN is the most complex system where there exist mixing of PSO optimization algorithm with BPNN activities. First weights are determined by using PSO, afterwards found weights are used in BPNN to test the system. Finally the results are presented and compared.

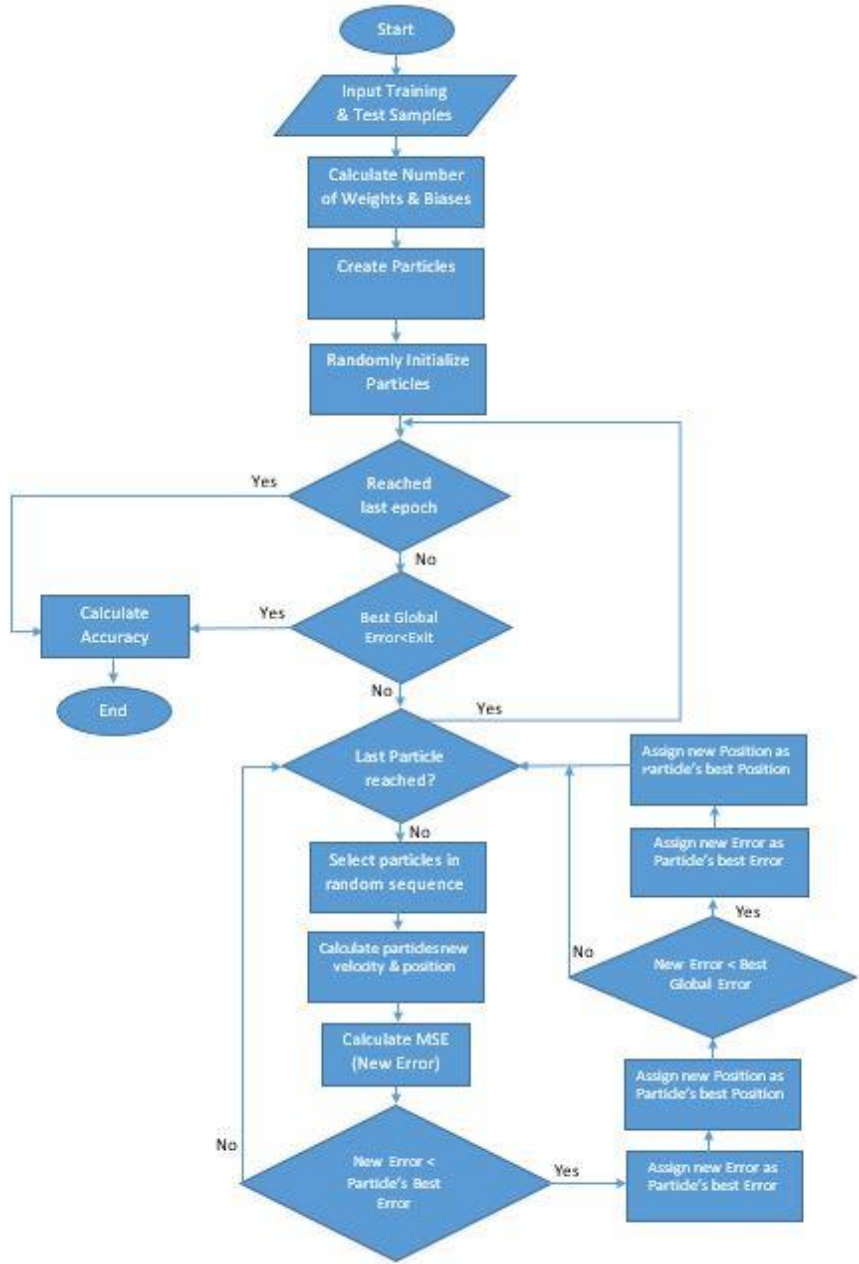

*Figure 1: The Proposed Flowchart of the PSOCNN Lecturers' Performance System*

The main procedure of the system is more described through the following pseudo-code of PSOCNN.

*Begin*
*Input: Training Samples; Testing Samples; Number of Input, Hidden and Output Neurons; Number of Particles; Max Epoches; Exit Error*
*Output: Accuracy of Train & Test Samples*
 *fetch Train and Test Samples*
 *Calculate Number of Weights using Number of Input, Hidden and Output Neurons*

```
  Create Particles
   for each Particle in the Swarm
    Randomly initialize Velocities and Positions (Weights & biases)
    Calculate fitness function error = MeanSquaredError(Train Samples,RandomPositions)
    if Particles error < Best Global Error then
      Assign error as besGlobalError
      Assign this particles Positions (Weights & biases) as besGlobalPosition
    endif
  endfor

while (epoch < Max Epoches)
 if bestGlobalError < Exit Error then break;
 endif
   for each Particle in the Swarm
    Select a Particle in Random Squence
    Calculate Particle's new Velocities using equation (9)
    Calculate Particle's new Position using new Velocity in equation (8)
    Calculate Particle's new error = MeanSquaredError(Train Samples, NewPositions(New weights & biases))
       if newerror < Particle's bestError then
          Assign newerror as Particle's bestError
          Assign newPosition (New weights & biases) as Particle's bestPosition (Best weights & biases)
       endif
     if newerror < bestGlobalError then
      Assign newerror as bestGlobalError
      Assign this Particle's newPosition (New weights & biases) as bestGlobalPosions(best weights & biases found ever)
     endif
   endfor
    epoch++
endwhile
 Calculate & Return Accuracy of both Train Samples & Test Samples
 MeanSquaredError (Train Samples, Positions(Weights & biases) ) (∗ Method to find fitness function ∗)
 Set weights & biases found as Positions of Particle
 for each Train Samples
  Compute predicted output
   sumSquaredError += ( predicted output – target output ) ^ 2
 endfor
   return (sumSquaredError / Number of Train Samples ) (∗ End of MeanSquaredError ∗)
   Accuracy (Samples) (∗ Method to find Accuracy of Train and Test Samples ∗)
  Set Weights found as besGlobalPositions (Best weights & biases found)
 for each Sample in the Sample Set
  Compute predicted output
    if predicted output differs by ±0.1 from target output then
     Apply round down  & round up to predicted output accordingly
     number_of_correctly_classified++
   else
     Apply normal rounding to predicted output
     number_of_incorrectly_classified++
   endif
 endfor
return (number_of_correctly_classified / number_of_correctly_classified + number_of_incorrectly_classified )
End
```

## 4. METHODOLOGY

The proposed lecturers' performance enhancement system using PSOCNN is implemented via four main steps; collecting data, pre-processing, feature selection, and classification. Initially, the data is collected from the college departments of SUE, then manual initial pre-processing is carried out to change collected data which are in hard copy formats and in excel sheets to a format to be suitable for the processing, and thus, we can call the final set of the initial pre-processed data database. The database is collected in three different subsets each of which is described in the next sub sections. Afterwards, samples of each subset are divided into two subsets called training and testing sets. The next step is pre-processing to perform some sort of balancing data in each class samples. Feature selection is then performed when all the three data sets are used together. Eventually, the final set is fed into PSOCNN. Detail of the stages will be described in the following sections:-

## 4.1 Performance Database

Data resources may be in-house or can be outsourced and since this research work is designed as an early warning system for the academic staffs at SUE. Thus, this work will use in-house data existing in the department of quality of assurance at SUE. The dataset is collected for the academic years 2011 and 2012. The data is analyzed, then, it has been processed to be suitable for processing. The data has three sections namely; Student Feedback, Lecturers Portfolio and Continuous Academic Development (CAD). In the next subsections details of the collected data are described:-

### 4.1.1 Student Feedback Sub-Dataset

One of the criteria that SUE uses for performance analysis is student feedback. This criterion shows the level of student acceptance of the lecturer according to teaching capability and behavior of the lecturer in the lecture class during one course of studying. Each student will give his/her feedback to the lecturer of the subject. Lecturers may have different subjects in the same department or in different departments, or in different colleges as well. The Sub-Dataset is composed of 12 features and 620 samples. The number of the samples of the student feedback

Sub-Dataset is different and is more than the other Sub-Dataset samples because a lecturer may be responsible for 2 or more subjects inside the university. The students are allowed to choose one option from a five rating scale measures to each of the above features and at the end the averages of the features are taken. The overall average is then labeled as A*, A, B, C, D. The labeling are used in the end with the other Sub-Dataset final labels to reach to a final result whether to give the corresponding lecturer award, warning or punishment.

### 4.1.2   Lecturers' Portfolio Sub-Dataset

The second criterion of measuring lecturer performance in the SUE is lecturer portfolio for one full academic year. The criterion shows the activity and the hardworking spirit of the lecturer in helping the quality of the education and university. The portfolio Sub-Dataset is composed of 11 features and 313 instances. Each college in SUE is responsible in establishing a committee of quality assurance for each department in order to evaluate lecturers' portfolio. The committee will consist of a number of lecturers and head of the department in the corresponding department.

### 4.1.3   Continuous Academic Development Sub-Dataset

The last criterion of measuring performance of lecturers in SUE University is continuous academic development. The criterion describes the activity and participation of lecturers in workshops seminars, committees and publications. The lecturers are given points according to their activities in the continuous learning process. The Sub-Dataset contains 3 features and 341 instances.

### 4.1.4   Combined Dataset

The combination of the above Sub-Datasets is prepared for the proposed system when all the Sub-Datasets are used as a single dataset as input to the system. The data set contains 313 instances with 26 features. The detail of the dataset is proposed in Table 1.

The data labeling for the final decision is released according to a special rule which is released from the quality assurance department in the university. The detail of the labeling of the final decision is described in Table 2.

### 4.2 Pre-Processing

As the data is collected from spread sheets and in hard copies, the data has to be transferred to a suitable format for the system. After transferring all the data to a soft copy, there were some missing values in some samples and the missing values where manually filled. The data are normalized in the pre-processing step to arrange all the data in the range between [0, 1].

### 4.3 Feature Selection

Detecting the relevant features will increase the accuracy rate of the system since it neglects the features with less relation to the system. Correlation-based Feature Subset Selection (CFS) is one of the techniques that is used for feature selection. Basically, CFS assesses the value of attributes in a subset through each feature's distinct analytical capacity alongside the redundancy degree between them [13, 14]. Features' subsets that have a higher relation with the class and are less correlated internally are preferred. The CFS technique is applied to the combined dataset. The dataset originally was composed of 26 features as it is described in Table 1. After applying the technique, the features with less relation to the class labels have been eliminated. The parameters of the technique when applying were left as default parameters. The final set of features after applying the technique has been reduced to 14 features. The final set of features is described in more detail in Table 3.

## 4.4 Classification and Experimental Results

In this section the system is explained in two different models (See Figure 2). Both models are compared to determine the one with the best accuracy. The number of samples are increased to 580 and used in both models. The original samples were 313 samples. 90% of the dataset is used for training and the remaining 10% is used to test the system. The PSO is used to find the best weights and biases in the training session of BPNN, afterwards those weights are used to test BPNN and results are compared. Detail descriptions of each of the models such as classifier parameters, test results, confusion matrices, and input data are described in the following subsections:-

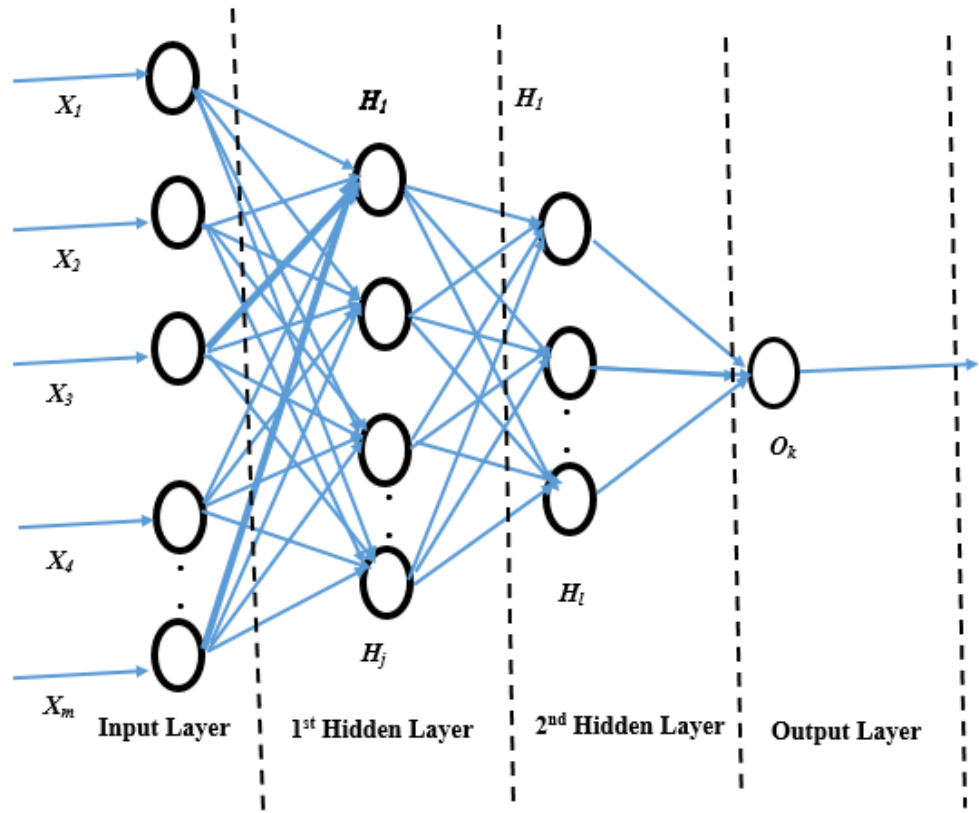

*Figure 2: The Structure of Both Models of PSOCNN*

### 4.4.1 Model 1

In this model, the preprocessed dataset with CFS is used as input features to the system. Then normalization is applied in the pre-processing to the input data. This model PSOCNN is used as classifier with PSO to find the weights and biases. The first model has 2 hidden layers (the first hidden layer with 12 neurons and the second hidden layer with 8 neurons) and it has 14 input neurons with only one output neuron. The parameters of first model classifier are presented in Table 4. The same parameters are used in both training and testing phases. Table 5 presents the results of both training and testing phases after applying the above parameter values to the network.

The matrix identifies that the misclassifications are very similar to each other. It shows that the misclassifications occur generally one class above or one class behind the actual class. The main reason is due to the similarity of the labeling function. The confusion matrix of the test phase of the first model which includes detail of the result (See Table 6). It found that testing phase of first model has the same properties with the confusion matrix of training phase where all the sample misclassifications are one class above or behind the actual class label.

### 4.4.2 Model 2

In order to reach a better accuracy this model is developed. The results in the first model show that misclassifications are due to only ±0.1 in the predicted output value where makes one class behind or over the actual output. The same dataset which is used in model 1, is again used with the same features and parameters. The only modification in this model is changing the comparison method of actual and predicted output by checking whether the difference is just ±0.1 or it is greater than this value. If it is greater than ±0.1 then normal rounding is used but if it is only ±0.1 then ceiling or flooring is used accordingly. Again PSO is used as weight and biases optimizer and BPNN is used as the main classifier. The parameters are shown in Table 7. After feeding the input data and applying the above parameters the results of both training and testing phases are presented in Table 8.

It is obviously seen that the misclassifications that were resulted due to a difference of ±0.1 is gone and now they are correctly classified. Again the reason for other misclassifications are due to the similarity of the labeling function. Table 9 shows detail of confusion matrix of the test phase of the second model. Again the modification had a great impact in the test phase as well as the misclassifications that were exist in the first model are correctly classified. As is it seen in the table, the second model has better accuracy than the first model which means the modification had its impact in the accuracy improvement.

The results indicate the outperformance of the second model against the first model in both testing and training phases (See Figure 3). The main reason is due to PSO algorithm that finds best weights for the network, these weights make the network become generalized. Results of the system shows that Model 2 outperforms Model 1, this is because of the modification that is done to the accuracy calculation that checks for ±0.1 difference between predicted and actual class which uses floor or ceiling functions accordingly. The modification is just useful for classification while it will not affect the result when the system is for prediction. Since this paper is based on classification, so, it is very useful to use this type of modification. The best model in the system has accuracy of 98.28 % of the testing phase which outperforms normal BPNN models without PSO.

## 5. CONCLUSION AND FUTURE RECOMMENDATION

In this paper, the following points are mainly concluded:
1) Combing data sets into one set will produce less accumulated errors which in return will cause the system performance and accuracy to increase.
2) Adding extra one hidden layer to case models in the proposed system increases the accuracy of the system.
3) Modifying accuracy calculation method by adding ceiling and floor rounding functions plays a great role in the increasing the accuracy.
4) This paper uses the CFS technique to get rid of interfering features that do not have any impact on recognizing the class label.

5) The highest accuracy rate in the PSOCNN system using PSO is achieved. The outperformance of this technique is due to using PSO for weight optimization and modification in the accuracy calculation method.

Although the combined approach provides good results for performance recognition, there are still some ways that could be considered to improve this approach. Some of the most important key ideas and recommendations for future work are listed below:-

1) Using other natural inspired algorithms (e.g. Artificial Bee Colony and Cuckoo algorithms) with BPNN to reach better accuracy in the system which needs improvements.

2) Using other databases that are already balanced and the features are clearer with clear class labeling functions.

*Table 1: Combined dataset features*

| Features | Description |
|---|---|
| PRF1 | Does the lecturer sincerely serve university educational process and dedicate required time for developing the process? |
| PRF2 | Does the lecturer actively help head of department and deanery for managing works? And does he/she implement the instructions given above him and mandated task on time? |
| PRF3 | Does the lecturer give significance to quality of teaching and has he taken steps to make quality in education and quality assurance certain? And does he/she participate in workshops and practicing courses? |
| PRF4 | In the new academic year, has the lecturer prepared appropriate course books for the lessons that he/she takes and has he explained the aim and principles of the course book to the students? |
| PRF5 | Is the lecturer active in doing exams and finding the errors, making analysis, findings and showing the last results to students? |
| PRF6 | Does the lecturer countercheck the criticism style and accepts the other side's criticism? And doesn't he/she make any difference between the students? |
| PRF7 | Is the lecturer's role in developing and managing scientific activities at the department/college positive? |
| PRF8 | Is the lecturer presented a successful academic year in his/her specialist field? Is he/she considered as good to be taken as an example? |
| PRF9 | Is the lecturer active in participating conferences in Kurdistan, Iraq and international ones, and does the lecturer participate in workshops and practicing courses? |
| PRF10 | Is the lecturer active in publishing researches and papers in domestic, Iraq and international journals? Has the lecturer's products been published in high graded international journals? |
| PRF11 | Has the lecturer participated in external occupational and charity activities that result in society interest? |
| CAD1 | Participating in seminars |
| CAD2 | Activities (Presenting seminars, publications, committee membership) |
| CAD3 | Summation of CAD1 + CAD2 |
| FB1 | Were the aim and the message of the course clear? |
| FB2 | Was the contents of the subject useful? Was it relevant to the course's main goals? |
| FB3 | Was the lecturer tried preparing course book? |
| FB4 | Did the lecturer try to explain principle, content and important topics during lessons easily? |
| FB5 | Did the lecturer start lessons on time and finish lessons on time? |
| FB6 | Did the lecturer behave quietly and respectfully during lessons? |
| FB7 | Were the used slides clear and attractive? |
| FB8 | Did the lecturer reserved a portion of time for student questions and try to answer all raised questions? |
| FB9 | Did the lecturer take the criticism of students into consideration? |
| FB10 | Was information about exams given? |
| FB11 | Were the questions of exams relevant to the content of the subject? |
| FB12 | Were the references new and suitable for contents of the subject? |

*Table 2: Final output decision labeling functions*

| CAD | FB | Portfolio | Label | Description |
|---|---|---|---|---|
| A* | A* | A | A | Thanks from Minister |
| A* or A | A* or A | B | B | Thanks from the Dean of the College |
| <=B | <=B | B or C | C | Rights remain the same |
| >=A | >=A | D | C | Rights remain the same |
| <=B | <=B | D | D | Warning |
| >=A | >=A | E | D | Warning |
| <=B | <=B | E | E | Firm Warning |

*Table 3: Combined dataset features after applying CFS*

| Features | Description |
|---|---|
| PRF1 | Does the lecturer sincerely serve university educational process and dedicate required time for developing the process? |
| PRF2 | Does the lecturer actively help head of department and deanery for managing works? And does he/she implement the instructions given above him and mandated task on time? |
| PRF3 | Does the lecturer give significance to quality and did he has taken steps to make quality in education and quality assurance certain, and does he/she participate in workshops and practicing courses? |
| PRF4 | in the new academic year, has the lecturer prepared appropriate course books for the lessons that he/she takes and has he has explained the aim and principles of the course book to the students? |
| PRF6 | Does the lecturer countercheck the criticism style and accepts the other side's criticism? And doesn't he/she make any difference between the students? |
| PRF8 | Is the lecturer presented a successful academic in his/her specialist field? Is he/she considered as good to be taken as an example? |
| PRF10 | Is the lecturer active in publishing researches and papers in domestic, Iraq and international journals? Has the lecturer's products |
| | been published in high graded international journals? |
| PRF11 | Is the lecturer participating in external occupational and charity activities that result in society interest? |
| CAD1 | Participating in seminars |
| CAD2 | Activities (Presenting seminars, publications, committee membership) |
| CAD3 | Summation of CAD1 + CAD2 |
| FB4 | Was the lecturer tried to explain principle, content and important topics during lessons easily? |
| FB5 | Did the lecturer start lessons on time and finish lessons on time? |
| FB10 | Was information about exams given? |

*Table 4: First Model – APSOCNN Classifier Parameters*

| Input Features | Output (Class) | Hidden | w | C1 | C2 | particles | Epoch |
|---|---|---|---|---|---|---|---|
| PRF1-PRF4, PRF6,PRF8, PRF10,PRF11, CAD1-CAD3, FB4,FB5,FB10 | 1,2,3,4,5 | 12,8 | 0.729 | 1.4944 | 1.4944 | 24 | 500 |

*Table 5: First Model - Results of Training and Testing Phase*

| Phase | CCNI | CCI (%) | ICNI | ICI (%) | MAE | RMSE |
|---|---|---|---|---|---|---|
| Training | 387 | 73.95 | 135 | 26.05 | 0.3940 | 0.4766 |
| Testing | 47 | 81.03 | 11 | 18.97 | 0.3805 | 0.4944 |

*Table 6: First Model - Confusion Matrix of Testing Phase*

| Classified As | | Thanks from Minister | Thanks from College Dean | Same Rights | Warning | Firm Warning |
|---|---|---|---|---|---|---|
| | | A | B | C | D | E |
| Thanks from Minister | A | 11 | 2 | 0 | 0 | 0 |
| Thanks from College Dean | B | 0 | 8 | 1 | 0 | 0 |
| Same Rights | C | 0 | 6 | 6 | 0 | 0 |
| Warning | D | 0 | 0 | 0 | 11 | 1 |
| Firm Warning | E | 0 | 0 | 1 | 0 | 11 |

*Table 7: Second Model – APSOCNN Classifier Parameters*

| Input Features | Output (Class) | Hidden | W | C1 | C2 | particles | Epoch |
|---|---|---|---|---|---|---|---|
| PRF1-PRF4, PRF6,PRF8, PRF10,PRF11, CAD1-CAD3, FB4,FB5,FB10 | 1,2,3,4,5 | 12,8 | 0.729 | 1.4944 | 1.4944 | 24 | 500 |

*Table 8: Second Model - Results of Training and Testing Phase*

| Phase | CCNI | CCI (%) | ICNI | ICI (%) | MAE | RMSE |
|---|---|---|---|---|---|---|
| Training | 509 | 97.51 | 13 | 2.49 | 0.0418 | 0.2837 |
| Testing | 57 | 98.28 | 1 | 1.72 | 0.0516 | 0.3939 |

Table 9: Second Model - Confusion Matrix of Testing Phase

| Classified As | | Thanks from Minister | Thanks from College Dean | Same Rights | Warning | Firm Warning |
|---|---|---|---|---|---|---|
| | | A | B | C | D | E |
| Thanks from Minister | A | 13 | 0 | 0 | 0 | 0 |
| Thanks from College Dean | B | 0 | 9 | 0 | 0 | 0 |
| Same Rights | C | 0 | 0 | 12 | 0 | 0 |
| Warning | D | 0 | 0 | 0 | 12 | 0 |
| Firm Warning | E | 0 | 1 | 0 | 0 | 11 |

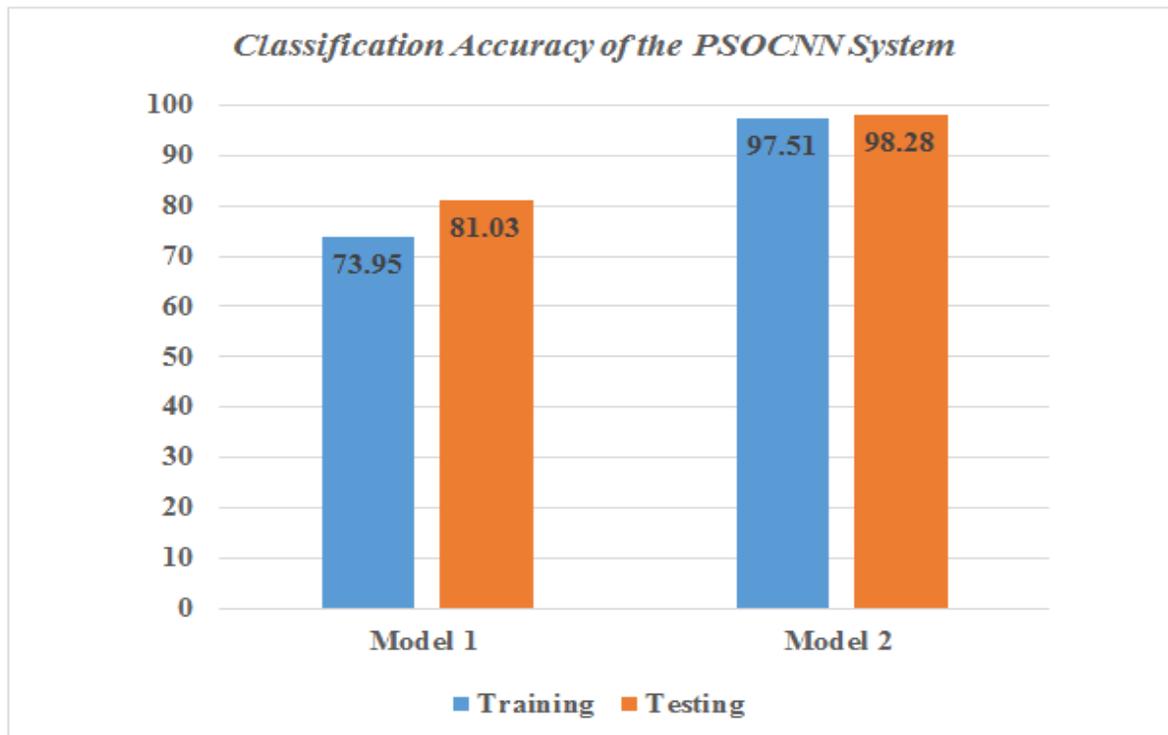

*Figure 3: The Percentage of Classification Accuracy of Both Models of the PSOCNN*